\documentclass[twocolumn]{article}
\usepackage[final]{nips_2017}
\usepackage[utf8]{inputenc} 
\usepackage[T1]{fontenc}    
\usepackage{hyperref}       
\usepackage{url}            
\usepackage{booktabs}       
\usepackage{amsfonts}       
\usepackage{nicefrac}       
\usepackage{microtype}      
\usepackage{graphicx}
\usepackage{amsmath}
\usepackage{subcaption}
\usepackage{wrapfig}
\usepackage{abstract}
\usepackage{tabularx}
\usepackage{float}
\usepackage[numbib]{tocbibind}
\title{Bridging the Gap Between Natural Language and Market Dynamics via High-Dimensional Representation Learning}

\author{
  Yujin Jeong
  \texttt{yujinjng@stanford.edu} \\
  \And
  Noelle Jung
  \texttt{noellej@stanford.edu} \\
  \AND
  Brian Y. C. Leung (Mike)
  \texttt{mbl@stanford.edu} \\
}

\makeatletter
\newcommand{\printnipsmaketitle}{\@maketitle}
\makeatother

\begin{document}

\twocolumn[{
  \begin{@twocolumnfalse}
    \begin{center}
        \includegraphics[width=3cm, height=0.7cm]{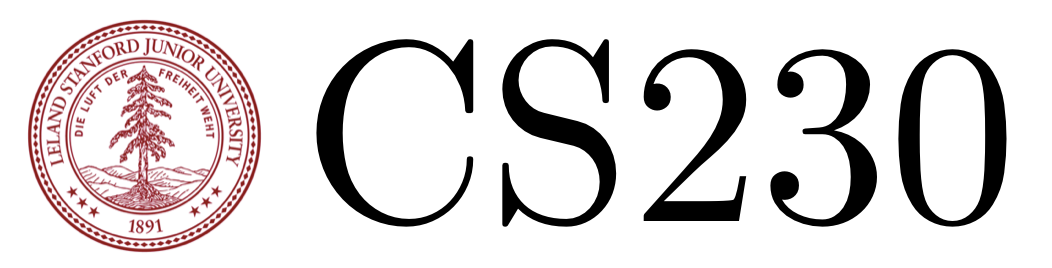}
    \end{center}
    
    \printnipsmaketitle
    \vspace{-20pt}
    \begin{abstract}
Traditional multi-modal financial forecasting often relies on scalar sentiment scores, which fail to capture the nuances of financial news. To address this information loss, this paper explores high-dimensional representation learning by replacing discrete polarity ratings with dense FinBERT embeddings within a Transformer-based forecasting architecture. We benchmarked various embedding strategies on the FNSPID dataset, including raw embeddings, attention-weighted aggregation, and a custom Siamese network. While the attention-based mechanism struggled with the low signal-to-noise ratio typical of financial data, the integration of Siamese-optimized embeddings outperformed both the scalar baseline and raw embedding approaches, demonstrating that preserving high-dimensional narrative context yields improved predictive accuracy for short-term stock price movements.
    \end{abstract}
    
    \vspace{1cm} 
  \end{@twocolumnfalse}
}]

\section{Introduction}	
Recent studies \cite{jun_Gu_2024}\cite{10796670} demonstrate the utility of capturing financial sentiment for financial modeling. Motivated by the shift towards high-dimensional representation learning \cite{vinden2025contrastivesimilaritylearningmarket}\cite{guo2024finetuninglargelanguagemodels}, this project aims to bridge the gap between sentiment classification and direct market prediction. We first replicate established baselines \cite{dong2024fnspid} that fuse structured data with prompt-generated sentiment scores, then extend this work by replacing scalar sentiment with richer semantic representations. Specifically, we investigate leveraging raw FinBERT \cite{Huang2023FinBERTAL} embeddings,  unfreezing FinBERT for stock prediction, training a custom Siamese network to learn a task-specific embedding, and using attention to weight sentiment averages. The input to our algorithm is a sequence of 50 days of financial news embeddings and structured price data. We then use a Transformer-based architecture to output a predicted closing price 3 days in the future. Our code can be found on Github \cite{Github}.

\section{Related work}
Past research in multi-modal stock prediction that we examined can be grouped into (1) scalar sentiment integration and (2) semantic representation.

\textit{\textbf{Scalar Sentiment Integration.}} This traditional approach reduces text to a polarity score prior to fusing with structured data. We draw on Dong et al. \cite{dong2024fnspid} who established baselines on the \textbf{Financial News and Stock Price Integration Dataset (FNSPID)} by combining ChatGPT-derived discrete sentiment ratings with Transformer architectures. Similarly, Gu et al. \cite{jun_Gu_2024} proposed the FinBERT-LSTM model, which uses FinBERT to generate sentiment indicators that are fed into an LSTM alongside historical prices. While computationally efficient and interpretable, we posit that this method suffers from significant information loss, and that projecting nuanced financial narratives into scalars discards the context required for detecting market signals.

\textit{\textbf{Semantic Representation and Direct Prediction.}} Recent state-of-the-art research shifts toward mapping dense text embeddings directly to market movements. Works like \cite{vinden2025contrastivesimilaritylearningmarket} and \cite{guo2024finetuninglargelanguagemodels} apply contrastive learning and fine-tuning of general LLMs (e.g., Mistral, LLaMA) to map text embeddings directly to market movements, capturing task-specific semantic nuances. Our experiments align with this approach but differ by utilizing article summaries rather than headlines to capture richer context. We also bring together FinBERT \cite{Huang2023FinBERTAL} and latent semantic analysis (LSA) summarization \cite{gong2001generic} to the FNSPID dataset to set a new useful benchmark. Unlike generic models, FinBERT is pre-trained specifically on financial texts to better grasp domain-specific contexts. LSA employs singular value decomposition to extract semantically dense sentences, balancing input length against information-density.

\section{Dataset and Features}
We used FNSPID \cite{dong2024fnspid}, selecting Google (GOOG), Microsoft (MSFT), Nvidia (NVDA), Apple (AAPL), and Amazon (AMZN). These tickers were chosen to match the original selection criteria from \cite{dong2024fnspid}, which focused on the 5 most influential stocks in the S\&P 500.

\subsection{Dataset Construction for Market Prediction} FNSPID consists of daily stock metrics aligned with financial news summaries. We constructed a time-series corpus spanning the available daily price-news history for the selected tickers, resulting in a total of 37,707 records. The data was split chronologically to prevent data leakage, allocating the first 80\% for training and the most recent 20\% for validation, according to backtesting best practices. This resulted in 30,165 training examples and 7,542 validation examples.

\textit{\textbf{Feature Engineering and Normalization.}} Consistent with \cite{dong2024fnspid}, we utilized \textit{Open}, \textit{Close}, and \textit{Trade Volume} as structured inputs. To ensure numerical stability, these features were Min-Max normalized per stock and data partition.

\textit{\textbf{Sentiment Signal Processing.}} First we replaced the ChatGPT API used in \cite{dong2024fnspid} with FinBERT. For the scalar baseline, FinBERT probabilities ($P_{pos}$, $P_{neg}$, $P_{neu}$) were mapped to a discrete sentiment score $S$ in [1, 5] using the formula:
\begin{equation}
    S = (P_{neg} \times 1.0) + (P_{neu} \times 3.0) + (P_{pos} \times 5.0)
\end{equation}
Daily scores were computed by averaging all articles released on a given trading day. Data sparsity proved to be a challenge, with coverage of summaries for our selected stocks averaging only 7.3\%. To impute missing sentiment, we adopted the recursive decay mechanism used by \cite{dong2024fnspid}. This approach assumes market sentiment persists but decays toward neutrality ($S_{neutral}=3$) over time:
\begin{equation}
  S(t) = S_{neutral} + (S(0) - S_{neutral}) \cdot e^{-\lambda t}
\end{equation}
where $\lambda=0.03$ represents the decay rate.
For our later explorations on high-dimensional embeddings, we used forward-filling to impute missing embeddings.

For visualization of sentiment data, refer to Figures \ref{fig:sentimentcoverage} and \ref{fig:sentimentdistribution} in the Appendix.

\textit{\textbf{Time-Series Discretization.}} Following the methodology established in \cite{dong2024fnspid}, the data was arranged into sliding windows consisting of a lookback window of $L=50$ days to predict a target horizon of $H=3$ days into the future. Thus our final input and output dimensions become:

\begin{itemize}
    \item Input: $X \in \mathbb{R}^{B \times 50 \times F}$
    \item Output: $y \in \mathbb{R}^{B \times 1}$, representing the \textit{Close} price at $t+3$
\end{itemize}

where $B$ is the batch size and $F$ is the number of features. For our baseline replication, $B=64$ and $F=4$.

\subsubsection{Example data}

\begin{table}[h!]
\centering
\resizebox{\columnwidth}{!}{
    \begin{tabular}{ |c|c|c|c|c|c| } 
     \hline
     $t$ & Date & Open & Close & Volume & Avg. Sent. \\ 
     \hline
     0 & 2023-01-11 & 133.51 & 132.75 & 131.25M & 2.92 \\ 
     1 & 2023-01-12 & 134.26 & 132.67 & 133.88M & 2.89 \\ 
     ... & ... & ... & ... & ... & ... \\
     49 & 2023-03-23 & 161.55 & 158.29 & 158.83M & 3.12 \\
     \hline
    \end{tabular}
}
\caption{Example 50-day window for AAPL}
\label{table:aapl50}
\vspace{-0.5cm}
\end{table}

\begin{table}[h!]
\centering
\begin{tabularx}{\columnwidth}{ |c|c|X| } 
 \hline
 $t$ & Date & News Summary (LSA) \\ 
 \hline
 ... & ... & ... \\
 47 & 2023-03-21 & "Rising competition from TikTok, privacy policy changes from Apple, and the raging war..." \\ 
 47 & 2023-03-21 & "Apple (AAPL) has grown to become one of the most relevant and influential companies..". \\ 
 48 & 2023-03-22 & "Apple Inc AAPL.O and Nvidia Corp NVDA.O, up 0.4\% and 2.4\% respectively..." \\
 ... & ... & ... \\
 \hline
\end{tabularx}
\caption{Example news summaries for AAPL}
\label{table:aapl}
\vspace{-0.5cm}
\end{table}

\begin{table}[h!]
\centering
\begin{tabular}{ |c|c| } 
 \hline
 Target date & Target Close price \\
 \hline
 2023-03-28 ($t=52$) & 157.01 \\
 \hline
\end{tabular}
\caption{Target variable for Table \ref{table:aapl50}}
\vspace{-0.5cm}
\end{table}

\subsection{Dataset Construction for Representation Learning}
To train our custom Siamese network we curated a distinct subset of data from the FNSPID tables.

\textit{\textbf{Label Generation for Contrastive Learning.}} Our Siamese network requires pairs of news items labeled by their market impact. We defined the ground-truth market movement $y_{market}$ as the forward close-to-open return, calculated as the change from the previous day's close to the next day's open:
\begin{equation}
    y_{market} = \frac{\text{Open}_{t+1} - \text{Close}_{t-1}}{\text{Close}_{t-1}}
\end{equation}
We selected the previous day's close ($\text{Close}_{t-1}$) as the baseline to ensure the reference price is strictly prior to any information released on Day $t$ to avoid lookahead bias. We filtered outliers using an Isolation Forest \cite{liu2008isolation} (5\% contamination) and removed extreme price movements (below 1st and above 99th percentiles) to ensure stability, leaving 1,991,435 rows of data. Refer to Figure \ref{fig:siamesehistogram} in the Appendix for a distribution of $y_{market}$ values.

\textit{\textbf{Discretization.}} To generate positive and negative pairs, we discretized the continuous $y_{market}$ values into bins. We experimented with three binning strategies: (1) Quartile, (2) Median and (3) Tercile.

\textit{\textbf{Sampling.}} We sampled 15,000 distinct price-news pairs, split into 12,000 for training/validation and a held-out set of 3,000 for testing the embedding quality across all three bin strategies. We excluded news from the 5 stocks used for the downstream task as Siamese training pairs should be isolated from the end-to-end Transformer architecture validation set to prevent data leakage.

\section{ Methods }
Our research consists of three phases: replicating the scalar baseline, introducing raw embedding integration, and optimizing the embedding space via unfreezing FinBERT, Siamese networks, and attention-based aggregation.

\subsection{Benchmark Replication}
In this phase, we replicated both the Transformer and LSTM baselines. As \cite{dong2024fnspid} did not disclose specific stock symbols or hyperparameters, we approximated the experiment setup to generate comparable results. Consequently, our validation focused on matching relative performance trends rather than exact numerical values. While we observed the expected performance gains when adding FinBERT sentiment scores to the Transformer, the LSTM model displayed mixed results rather than consistently benefiting from the sentiment data. This aligns with the findings of \cite{dong2024fnspid}. From this point onward, we focus on the Transformer model since \cite{dong2024fnspid} highlighted it as the best performing across six tested architectures.

\subsection{Replacing Sentiment Score with Embedding}
We hypothesized that both ChatGPT and FinBERT's final projection layer remove valuable information when producing a sentiment score. This motivated us to remove the projection layer and feed the FinBERT embedding directly into the training and inference input. Validating this hypothesis was critical in determining whether to continue improving upon embeddings for the downstream task of stock prediction.

\subsection{Further Embedding Improvement}
Although FinBERT improves upon BERT for analyzing text in a financial context, it was trained on linguistic tasks rather than stock market forecasting. We hypothesized that inputting optimized FinBERT embeddings into our Transformer together with the structured data would lead to better predictions. We approached this task in 3 different ways: (1) unfreezing FinBERT layers, (2) applying contrastive learning, and (3) using attention weighted aggregation.

\subsubsection{Unfreezing FinBERT}
End-to-end training typically yields higher performance as the encoder learns to extract exactly the information downstream layers need.

In \cite{dong2024fnspid}'s baseline architecture, sentiment scores are precomputed and fed into the model’s input, prohibiting back propagation into the FinBERT layers. We prepared $(\textbf{news\_tokens}, open\_price, close\_price, trade\_volume)$ instead of $(\textbf{sentiment\_score}, open\_price, close\_price, trade\_volume)$ as inputs, so that FinBERT could be brought into the model “in-graph”.

\subsubsection{Contrastive Learning}
To align the semantic space of our text embeddings with financial outcomes, we implemented a Siamese network.

\textit{\textbf{Network Architecture.}} The architecture consists of two identical sub-networks (towers) with shared weights. The input to each tower is the 768-dimensional embedding vector generated by a pretrained FinBERT model. Rather than using these high-dimensional embeddings directly, we process them through a Multi-Layer Perceptron (MLP) to introduce non-linearity and to filter linguistic noise. The specific sequence of layers in each tower is: \textbf{Linear Layer} (768 $\rightarrow$ 256), \textbf{ReLU Activation}, \textbf{Linear Layer} (256 $\rightarrow$ 64). The output is a pair of 64-dimensional vectors representing the two input articles in the optimized manifold.
\begin{figure}[htbp]
\vspace{-0.3cm}
    \centering
    \includegraphics[width=0.9\linewidth]{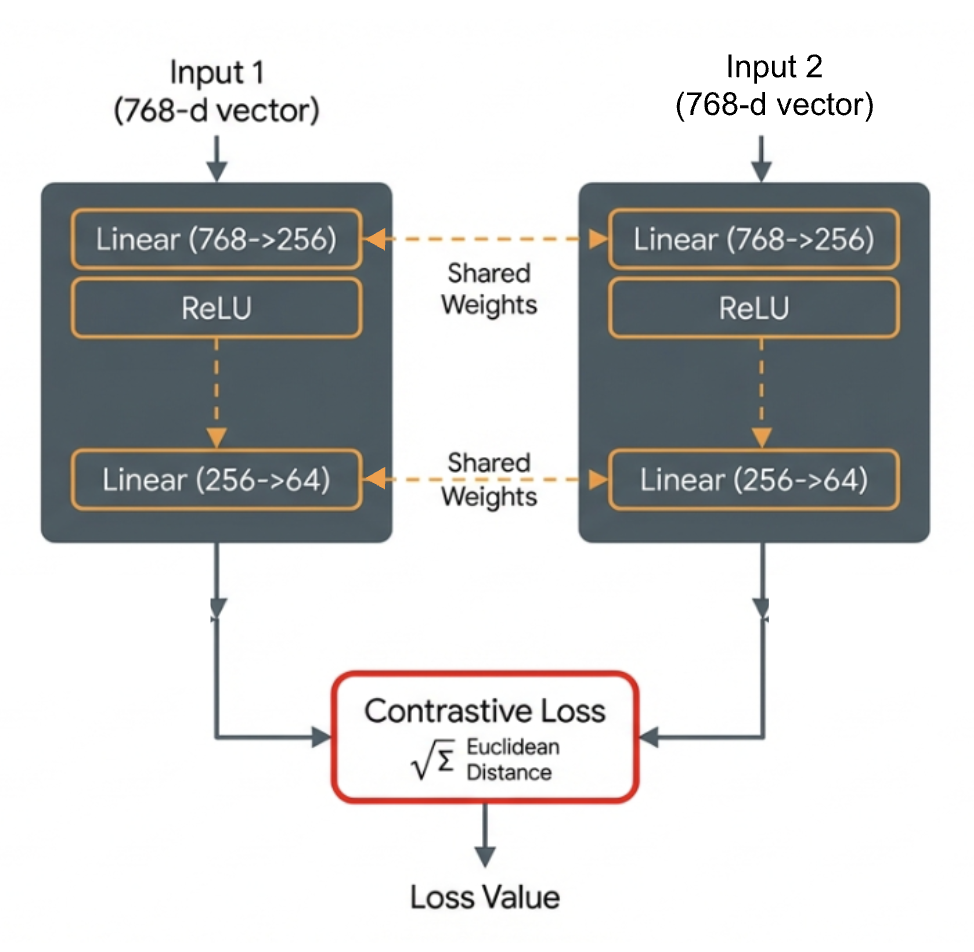}
    \caption{Diagram of Siamese network architecture}
    \label{fig:siamesediagram}
    \vspace{-0.5cm}
\end{figure}

\textit{\textbf{Quantile Binning.}} We constructed training pairs $(X_1, X_2)$ from our dataset using quantile binning on $y_{market}$. We labeled pairs as "similar" ($Y=1$) if they fell into the same bin and "dissimilar" ($Y=0$) otherwise. We chose to use quantile bins because financial data is inherently stochastic. We experimented with three binning strategies to determine the optimal method for constructing contrastive pairs: (1) \textbf{Quartile} (25/50/25), (2) \textbf{Median} (50/50) and (3) \textbf{Tercile} (33/33/33).

\textit{\textbf{Contrastive Loss.}} The network was trained using Contrastive Loss, which minimizes the Euclidean distance $D_w$ between positive pairs and maximizes the distance between negative pairs up to a margin $m=1.0$. The loss function is  defined as:
\begin{equation}
    L=\frac{1}{2}[Y \cdot D^2_w + (1-Y)\cdot \max{\{0,m-D_w\}}^2]
\end{equation}
By minimizing this loss, the shared encoder learns a metric space where proximity reflects market movement similarity.

\subsubsection{Attention-Based Daily Sentiment Aggregation}
The standard mean pooling approach, as used in \cite{dong2024fnspid}, is suboptimal due to significant variance in daily news volume, causing high-volume days to dilute high-signal narratives with noise. Refer to Figure \ref{fig:nvda} in the Appendix as an example of variance in daily news volume for Nvidia.

To mitigate this, we implemented a stock-specific attention mechanism. We formulate a Query-Key-Value operation where the \textbf{Query (Q)} is a learnable stock-specific vector, the \textbf{Key (K)} represents the article's semantic topic via SentenceBERT, and the \textbf{Value (V)} is the FinBERT sentiment embedding. We compute attention weights $\alpha_{s,i}$ via softmax on the dot-product of the stock query $q_s$ and article key $k_i$
:
\begin{equation}
    \alpha_{s,i} = \text{softmax}\left( \frac{q_s \cdot k_i}{\sqrt{d_k}} \right) = \frac{\exp(q_s \cdot k_i)}{\sum_{j=1}^{|N|} \exp(q_s \cdot k_j)}
\end{equation}
These weights then aggregate the FinBERT values into the daily sentiment embedding $E_{s,t}$:
\begin{equation}
    E_{s,t} = \sum_{i=1}^{|N|} \alpha_{s,i} v_i
\end{equation}
This process dynamically emphasizes articles whose topics align with the stock's learned profile.

\section{Experiments}

\begin{table}[h!]
\centering
\begin{tabular}{ |l|r|r|r| } 
 \hline
 LSTM-based Models & MSE & MAE & $R^2$ \\ 
 \hline
 LSTM Baseline (No Sentiment) & \textbf{0.172430} & \textbf{0.173834} & \textbf{0.931680} \\
 LSTM Baseline (Sentiment Score) & 0.223257 & 0.201129 & 0.911542 \\
 \hline
\end{tabular}
\caption{Performance of LSTM-based Baseline Models}
\label{table:lstm}
\vspace{-0.5cm}
\end{table}

\begin{table}[h!]
\centering
\resizebox{\columnwidth}{!}{
    \begin{tabular}{ |l|r|r|r| } 
     \hline
     Transformer-based Models & MSE & MAE & $R^2$ \\ 
     \hline
     Transformer Baseline (No Sentiment) & 0.244316 & 0.298697 & 0.903198 \\
     Transformer Baseline (Sentiment Score) & 0.199477 & 0.283591 & 0.920964 \\
     Transformer + Embeddings (Frozen FinBERT)  & 0.165027 & 0.225731 & 0.934614 \\
     Transformer + Unfreezing FinBERT (1 epoch) & 2.406112 & 0.867279 & 0.046657 \\
     Transformer + Siamese Network & \textbf{0.078109} & \textbf{0.144500} & \textbf{0.969052} \\
     Transformer + Attention Weighted Aggregation & 0.199007 & 0.253417 & 0.921150 \\
     \hline
    \end{tabular}
}
\caption{Performance of Transformer-based Baseline and Experiment Models}
\label{table:transformer}
\vspace{-0.5cm}
\end{table}

\subsection{Benchmark Replication}
We follow \cite{dong2024fnspid}'s evaluation framework to use MSE, MAE, and $R^2$ as our comparison yardstick. We selected the same hyperparameters used by \cite{dong2024fnspid} (100 epochs, 4 vanilla transformer layers). We utilize the Mean Squared Error (MSE) as our loss function, consistent with \cite{dong2024fnspid}, defined as:

\begin{equation}
    \mathcal{L}_{\text{MSE}} = \frac{1}{N} \sum_{i=1}^{N} (y_i - \hat{y}_i)^2
    \label{eq:mse_loss}
\end{equation}

\noindent where $N$ represents the number of samples in the batch, $y_i$ denotes the ground truth value, and $\hat{y}_i$ is the predicted value generated by the model. This configuration remains consistent across all replication and experimental models.

We made reasonable choices for parameters that were not explicitly published (e.g. default AdamW optimizer, 4 transformer heads, 0.2 for dropout, using a learning rate scheduler with a 30\% ramp-up, with a max learning rate of $1\times10^{-3}$) and kept these consistent across experiments. We found that these choices gave us more consistent loss improvement trends and helped prevent overfitting. We experimented with learning rates and found that exceedingly low values waste compute as both training and validation loss don’t change, while high values cause training loss fluctuation. We found that using a learning rate scheduler and 30\% ramp-up period helped us achieve a relatively consistent downward loss trend and generalized well for our subsequent experiments. Drop out, as expected, helped with overfitting to the training set. 

For normalization, \cite{dong2024fnspid} mentioned that they used Min-Max scaling, but did not specify exactly how. We tried normalizing stock prices globally and found that it led to a data leakage problem that resulted in inflated validation metrics, as the test data sees some information from the validation set. We tested different normalization configurations and settled on strictly splitting normalization between test and validation, and doing this per stock, using Min-Max scaling. Normalization was applied for targets as well.

We observed that adding FinBERT sentiment achieves the same directional effect as FNSPID’s benchmark (see Tables \ref{table:lstm} and \ref{table:transformer}). On Transformer-based models, we achieved a bigger magnitude of improvement across all 3 metrics.

\begin{table}[h!]
\centering
\resizebox{\columnwidth}{!}{
    \begin{tabular}{ |c||r|r|r| } 
     \hline
      & Transformer Baseline & Transformer Baseline &  \\ 
      & (No Sentiment) & (Sentiment Score) & Difference \\
     \hline
     MSE & 0.244 & 0.199 & -18.44\% \\
     MAE & 0.299 & 0.284 & -5.02\% \\
     $R^2$ & 0.903 & 0.921 & +1.99\% \\
     \hline
    \end{tabular}
}
\caption{Performance impact of sentiment score on Transformer baseline}
\vspace{-0.75cm}
\end{table}

\begin{table}[h!]
\centering
\begin{tabular}{ |c||r|r|r| } 
 \hline
  & LSTM Baseline & LSTM Baseline & \\ 
  & (No Sentiment) & (Sentiment Score) & Difference \\
 \hline
 MSE & 0.172 & 0.223 & +29.65\% \\
 MAE & 0.174 & 0.201 & +15.52\% \\
 $R^2$ & 0.932 & 0.912 & -2.15\% \\
 \hline
\end{tabular}
\caption{Performance impact of sentiment score on LSTM baseline}
\vspace{-1cm}
\end{table}

\subsection{Replacing Sentiment Score with embedding}
For this experiment we used the same hyperparameters as the benchmark replication in order to maintain consistency for a direct comparison of metrics.

The results of this experiment (see Table \ref{table:transf-emb}) validated our hypothesis that replacing sentiment score with a daily embedding, calculated as the average of the individual last-layer FinBERT article embeddings, would produce better predictions. Going forward, \textbf{“Transformer + Embeddings (Frozen FinBERT)”} serves as our iteration platform for experimental text embeddings for the task of stock price prediction.

\begin{table}[h!]
\centering
\resizebox{\columnwidth}{!}{
    \begin{tabular}{ |c||r|r|r| } 
     \hline
      & Transformer Baseline & Transformer + Embeddings & \\ 
      & (Sentiment Score) & (Frozen FinBERT) & Difference \\
     \hline
     MSE & 0.199 & 0.165 & -17.09\% \\
     MAE & 0.284 & 0.226 & -20.42\% \\
     $R^2$ & 0.921 & 0.935 & +1.52\% \\
     \hline
    \end{tabular}
}
\caption{Performance impact of news embeddings on Transformer model}
\label{table:transf-emb}
\vspace{-0.5cm}
\end{table}

\subsection{Further Embedding Improvement}

\subsubsection{Unfreezing FinBERT}
Our first step in attempting to improve the sentiment embedding was task-specific fine tuning. We discovered that fine tuning even one layer of FinBERT is expensive. Freezing FinBERT and only training 4 layers of vanilla transformers of 4 heads using representation dimensionality of 32 had 75k trainable parameters. Unfreezing just one layer of FinBERT would result in 7.7 million trainable parameters. With our data and model configuration, each epoch would take \~50 minutes to train on a T4 GPU. For fair comparison, each full iteration would take 100 hours. We finished training 1 epoch, but ultimately abandoned this direction and opted for computationally more efficient directions discussed below. 

\subsubsection{Contrastive Learning}
We trained the Siamese network using the ADAM optimizer with an initial learning rate of $1\times10^{-4}$. We chose this lower rate to keep the training process stable, as contrastive loss can be sensitive to large updates. We employed a batch size of 64, which fit within our memory limits. The model was trained for 15 epochs with a learning rate scheduler, which dynamically decayed the learning rate when validation loss plateaued, allowing fine-grained optimization in later epochs. 

We evaluated the three binning strategies on a common hold-out test set to ensure fair comparison. As shown in Figure \ref{fig:siamese_roc} in the Appendix, all three strategies resulted in ROC AUC scores near random chance (0.5), indicating that the Siamese network struggled to learn a robust manifold that linearly separates market reactions based solely on text embeddings. This is expected as each news summary merely represents a small part of the information that the market considers, so we only expect it to contribute marginally to the daily market movement. 

While the Quartile strategy achieved AUC comparable to Tercile (0.509 vs. 0.505), a deeper inspection of the confusion matrices revealed significant issues with model bias.

\begin{figure}[h]
  \centering 
  
  \begin{subfigure}{0.49\textwidth} 
    \centering
    \includegraphics[width=0.67\linewidth, height=0.5\linewidth, keepaspectratio=false]{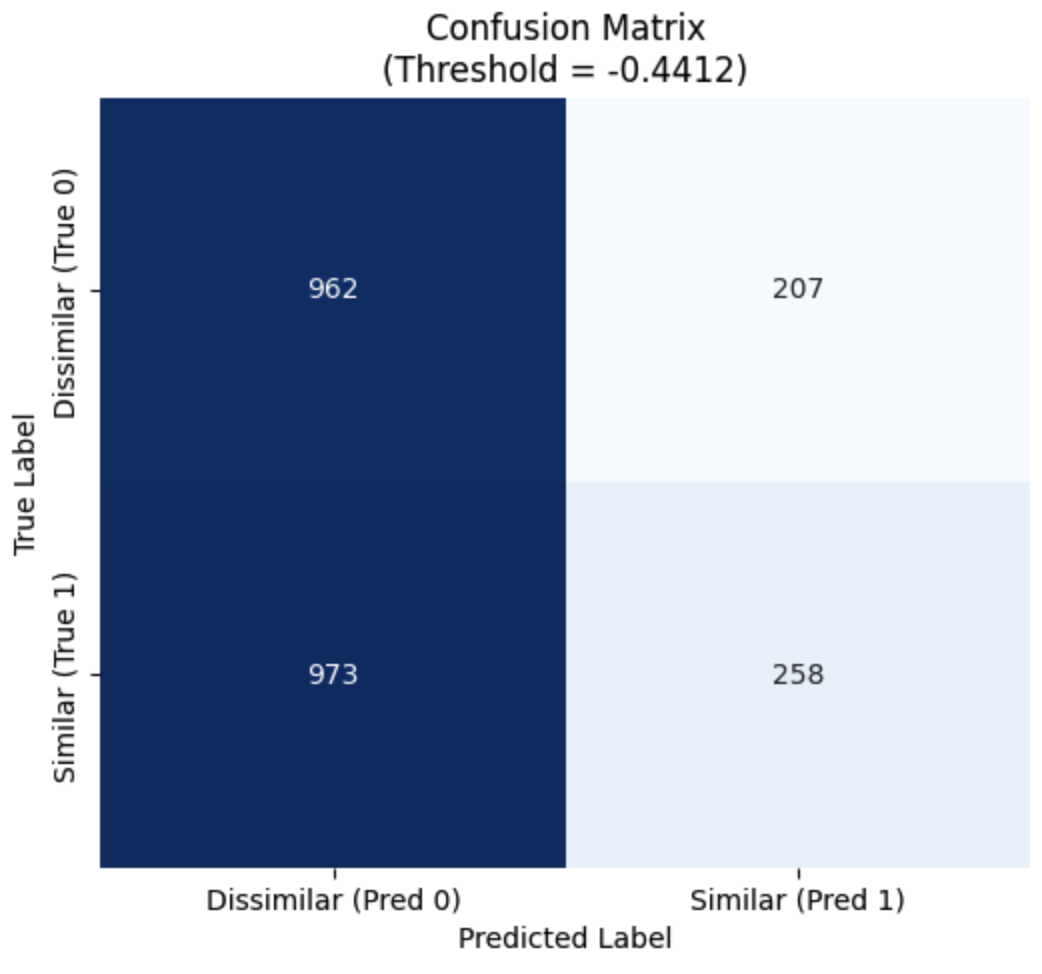} 
    \caption{Quartile 25/50/25 binning}
    \label{fig:quartilecm}
  \end{subfigure}
  \hfill 
  \begin{subfigure}{0.49\textwidth} 
    \centering
    \includegraphics[width=0.67\linewidth, height=0.5\linewidth, keepaspectratio=false]{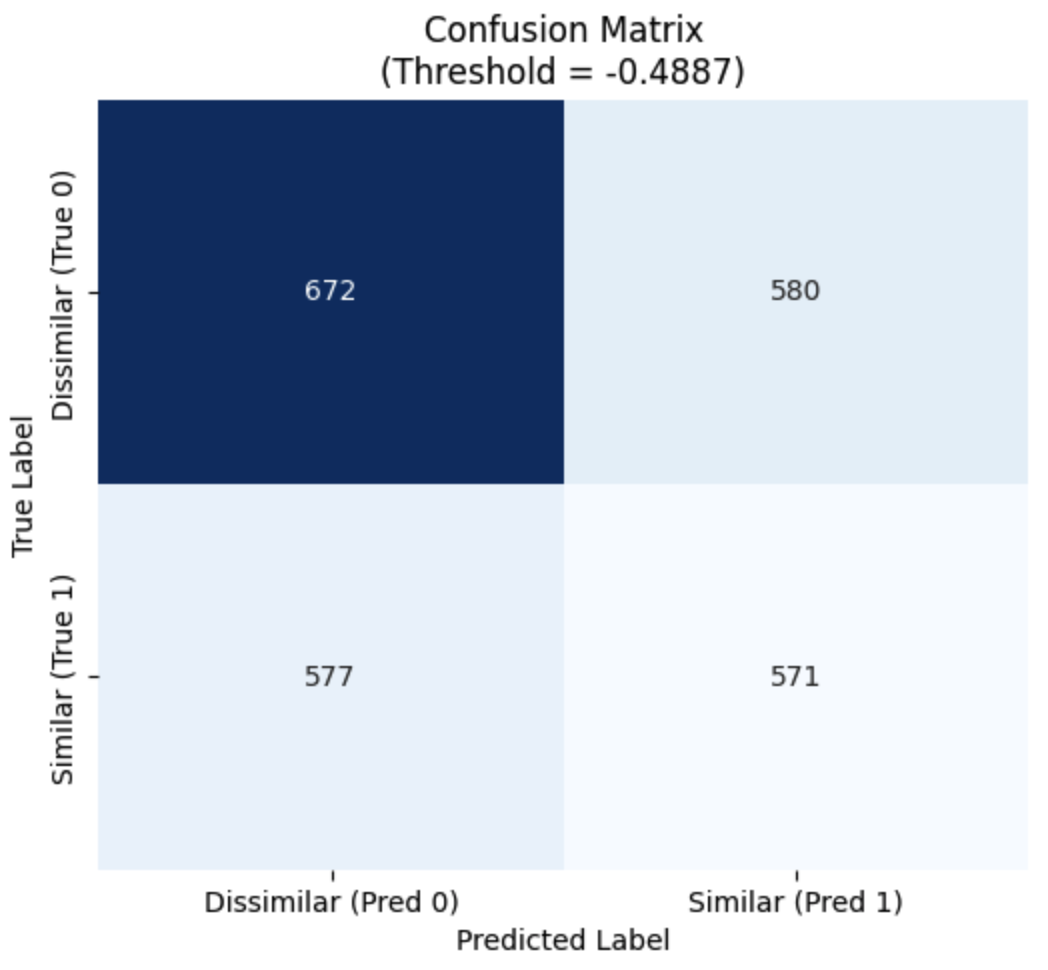}
    \caption{Tercile 33/33/33 binning}
    \label{fig:tercilecm}
  \end{subfigure}
  
  \caption{Confusion matrices for binning strategies}
  \label{fig:image2}
\end{figure}

The confusion matrix for the Quartile strategy (Figure \ref{fig:quartilecm}) shows a severe imbalance. The model correctly identified 962 dissimilar pairs (True Negatives) but only 258 similar pairs (True Positives), resulting in a Recall of just 20.9\%. This suggests the model collapsed into predicting "Dissimilar" for the majority of inputs. In contrast, the Tercile strategy demonstrated a much more balanced confusion matrix (Figure \ref{fig:tercilecm}). It achieved a Recall of 49.7\% and a Precision of 49.6\%. Although the overall AUC was slightly lower, this strategy forced the model to actually learn features distinguishing the two classes.

Loss curves for each Siamese network (see Figures \ref{fig:lossq}, \ref{fig:lossm}, and \ref{fig:losst} in the Appendix) show overfitting in Quartile and Median Split binning strategies. We addressed overfitting by tuning hyperparameters such as dropout, weight decay, and learning rate, but were not able to further reduce variance without creating an upward trend in validation loss. This is likely due to the inherent difficulty of the task as discussed above.

We chose the Tercile strategy for our final model, which seemed to be the most balanced and robust to model collapse. The Tercile strategy also showed slightly more stable validation loss during training, indicating it was less likely to memorize noise in the training data. This combination of better class balance and training stability made it the best candidate for generating the embeddings needed for our Transformer model.

\textit{\textbf{End-to-end Performance.}} The Transformer integrated with Siamese Network embeddings achieved the highest predictive performance metrics, reaching an MSE of 0.078109 and an $R^2$ of 0.969052 (see Table \ref{table:transformer}).

\subsubsection{Attention Weighted Aggregation }
For the attention mechanism, we utilized \textbf{Value Vectors (V):} 768-dimensional FinBERT embeddings for each article, \textbf{Key Vectors (K):} 384-dimensional SentenceBERT embeddings representing article topics, and \textbf{Query Vectors (Q):} Learnable 384-dimensional stock-specific vectors initialized randomly. Q dimension was chosen to enable dot-product calculation with K.

\begin{table}[h!]
\centering
\begin{tabular}{ |l||r|r|r| } 
 \hline
  Method & MSE & MAE & $R^2$ \\ 
 \hline
 Baseline (Mean Pooling) & 0.165 & 0.2257 & 0.9346 \\
 Attention Weighted Aggregation & 0.199 & 0.2534 & 0.9212 \\
 \hline
\end{tabular}
\caption{Performance Comparison of Aggregation Strategies (Horizon = 3 Days)}
\label{table:qkv}
\vspace{-0.65cm}
\end{table}

Table \ref{table:qkv} summarizes the performance of the Attention-Based Aggregation compared to the standard Mean Pooling baseline on the test set. Despite the theoretical appeal of filtering noise via attention, the experimental results suggest several factors contributed to its failure in this specific context.

Financial news datasets are characterized by a low Signal-to-Noise Ratio (SNR). As observed in similar domains \cite{jain2019attentionexplanation}, attention mechanisms trained on noisy, limited data often suffer from "attention collapse," where the learned weights $\alpha_{s,i}$ fail to distinguish meaningfully between high-value and low-value articles, effectively reverting to a (noisier) uniform average.

Introducing learnable queries and key-value projections added significant complexity to the optimization landscape. Without a dedicated auxiliary loss to guide the attention mechanism (e.g., a supervised signal on which articles were important), the model struggled to learn meaningful associations purely from future price movements.

\subsection{Tying Results to the Real World}
\begin{figure}[htbp]
    \centering
    \includegraphics[width=1.0\linewidth]{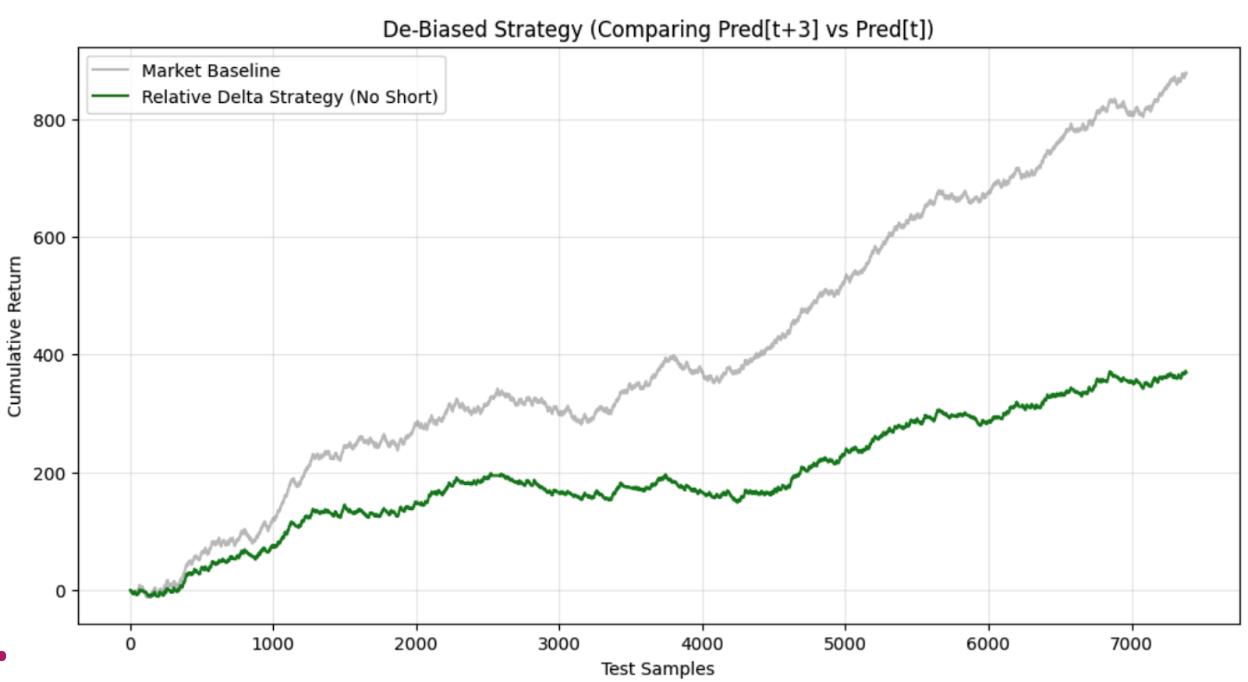}
    \caption{Comparison of sentiment-enhanced models vs. a passive buy-and-hold strategy}
    \label{fig:backtest}
\end{figure}

To translate our model's predictive performance into practical trading outcomes, we conducted a simple event-driven backtest. We initiated a hypothetical long position when the model predicted a positive price movement three days ahead; otherwise, we held cash to accrue the risk-free rate. The results were consistent across all models, including the baseline. With directional accuracies hovering near random chance (\textasciitilde50\%), none were able to outperform a passive buy-and-hold benchmark (see Figure \ref{fig:backtest}).

These results align with the Efficient Market Hypothesis \cite{emh}, suggesting that current asset prices largely reflect available information. Our findings demonstrate that even a sophisticated NLP-derived sentiment factor lacks the predictive power required to generate statistically significant alpha when used in isolation.

This does not imply that markets are perfectly efficient or that quantitative strategies are futile. Rather, it highlights the high bar for efficiency in modern markets. Single-factor signals are often quickly identified and arbitraged away, a phenomenon well-documented by McLean and Pontiff \cite{mclean}. In this competitive environment, edge is rarely found in isolation; instead, successful quantitative strategies rely on the robust aggregation of many weakly correlated factors \cite{valmoe}.

\section{Conclusion }
This study evaluated high-dimensional representation learning for financial forecasting, moving beyond scalar sentiment scores to preserve narrative nuance. Our benchmarks demonstrate that the Transformer integrated with Siamese Network embeddings outperformed all other architectures, including those using raw FinBERT embeddings and attention weighted aggregation. Although the Siamese network failed to effectively learn a "market-aware" geometric space, its embeddings combined with structured data improved predictions. Contrary to our hypothesis, the attention weighted average mechanism failed to improve performance. We attribute this to the low Signal-to-Noise Ratio (SNR) of financial news, which often leads to attention collapse, and the limitation of using static learnable query vectors to model dynamic market interests. A static query cannot adapt to the rapidly shifting narratives that drive stock prices over time.

Future research should prioritize addressing the Attention Collapse issue, potentially through the use of Differential Transformers\cite{ye2024differential} which use subtractive attention to cancel noise. Additionally, the unfreezing of FinBERT, deprioritized in this study due to compute constraints, remains a promising avenue for end-to-end optimization. Further exploration is needed to validate the model's predictive power and eliminate autocorrelation bias. Finally, the experimental scope should be expanded beyond the current five-ticker subset to the full FNSPID dataset to assess generalizability across diverse market sectors.

\section{Contributions}
All members contributed equally to literature review, idea selection, and writing reports.

Noelle focused on the LSTM baseline replication with and without sentiment scores and the contrastive learning experiment. The latter entailed creating scripts for processing news and price data, training the Siamese network, integrating the embeddings into the Transformer architecture, training the end-to-end model, and evaluating results.

Mike built initial prototypes of FinBERT Siamese network, trained a 1-day look-back variant of the baseline, trained the baseline variant that replaced sentiment scores with embeddings, trained finetuned FinBERT, and built the back testing code. He also set up Github and contributed to early data engineering and Colab pipeline setup. 

Yujin did data engineering and feature extraction using AWS Athena and SQL. She developed and refined early Siamese network experiments to formulate experiment ideas, conducted the benchmark replication of Transformer-based models, and implemented Attention-Based Daily Sentiment Aggregation.

\bibliographystyle{plain}
\bibliography{references}

\section{Appendix}
\label{app:data}
\subsection*{Implementation Details}

The experimental framework was primarily implemented using PyTorch~\cite{paszke2019pytorch} for deep learning model development, though initial comparative experiments were conducted using TensorFlow~\cite{abadi2016tensorflow}. 

For natural language processing tasks, we utilized the Hugging Face Transformers library~\cite{wolf2020transformers}. Data preprocessing, manipulation, and numerical computations were handled using Pandas~\cite{mckinney2010data} and NumPy~\cite{harris2020array}, while Scikit-learn~\cite{pedregosa2011scikit} was employed for evaluation metrics and baseline algorithms. 

Cloud data interactions were managed via the AWS SDK for Python (Boto3)~\cite{boto3} and AWS Data Wrangler~\cite{awswrangler}. Finally, all visualizations were generated using Matplotlib~\cite{hunter2007matplotlib}.

\subsection*{Charts and Figures}

\begin{figure}[htbp]
    \centering
    \includegraphics[width=1.0\linewidth]{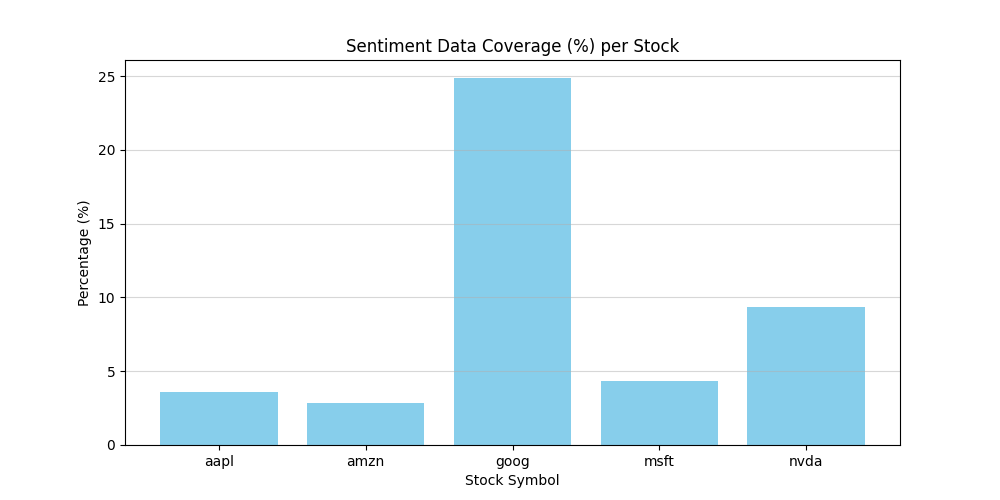}
    \caption{Percentage of sentiment data by stock symbol}
    \label{fig:sentimentcoverage}
\end{figure}

\begin{figure}[htbp]
    \centering
    \includegraphics[width=1.0\linewidth]{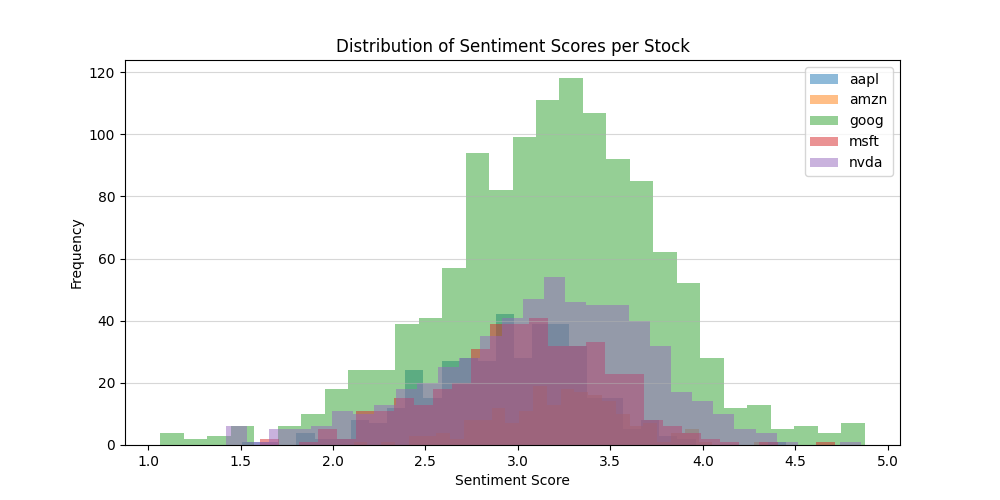}
    \caption{Percentage of sentiment data by stock symbol}
    \label{fig:sentimentdistribution}
\end{figure}

\begin{figure}[htbp]
    \centering
    \includegraphics[width=1.0\linewidth]{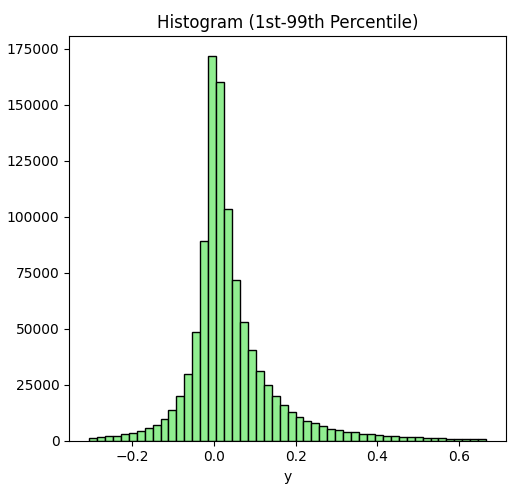}
    \caption{Distribution of values for $y_{market}$ in contrastive learning data}
    \label{fig:siamesehistogram}
\end{figure}

\begin{figure}[htbp]
    \centering
    \includegraphics[width=1.0\linewidth]{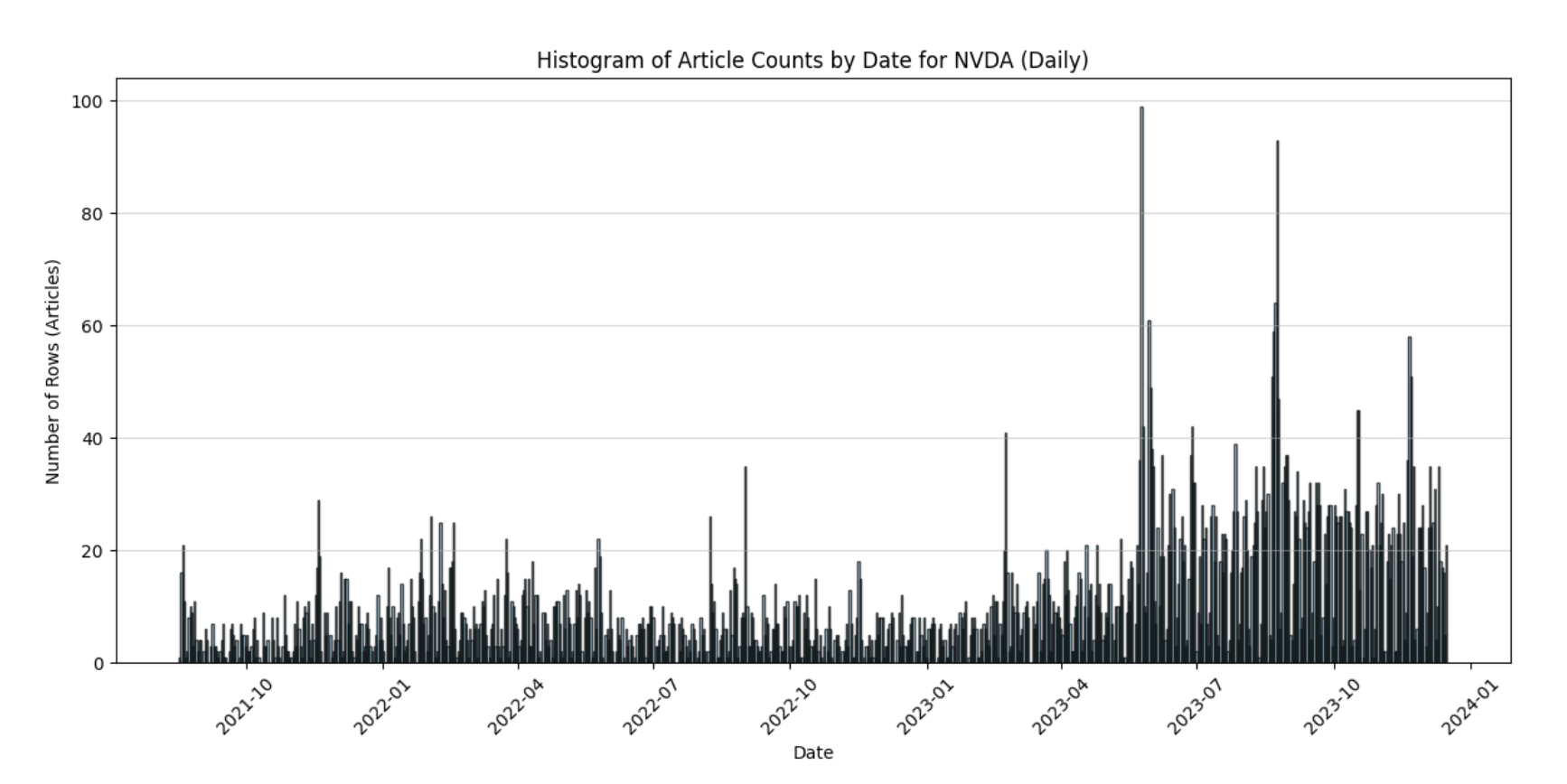}
    \caption{Article counts by date for NVDA}
    \label{fig:nvda}
\end{figure}

\begin{figure}[htbp]
    \centering
    \includegraphics[width=1.0\linewidth]{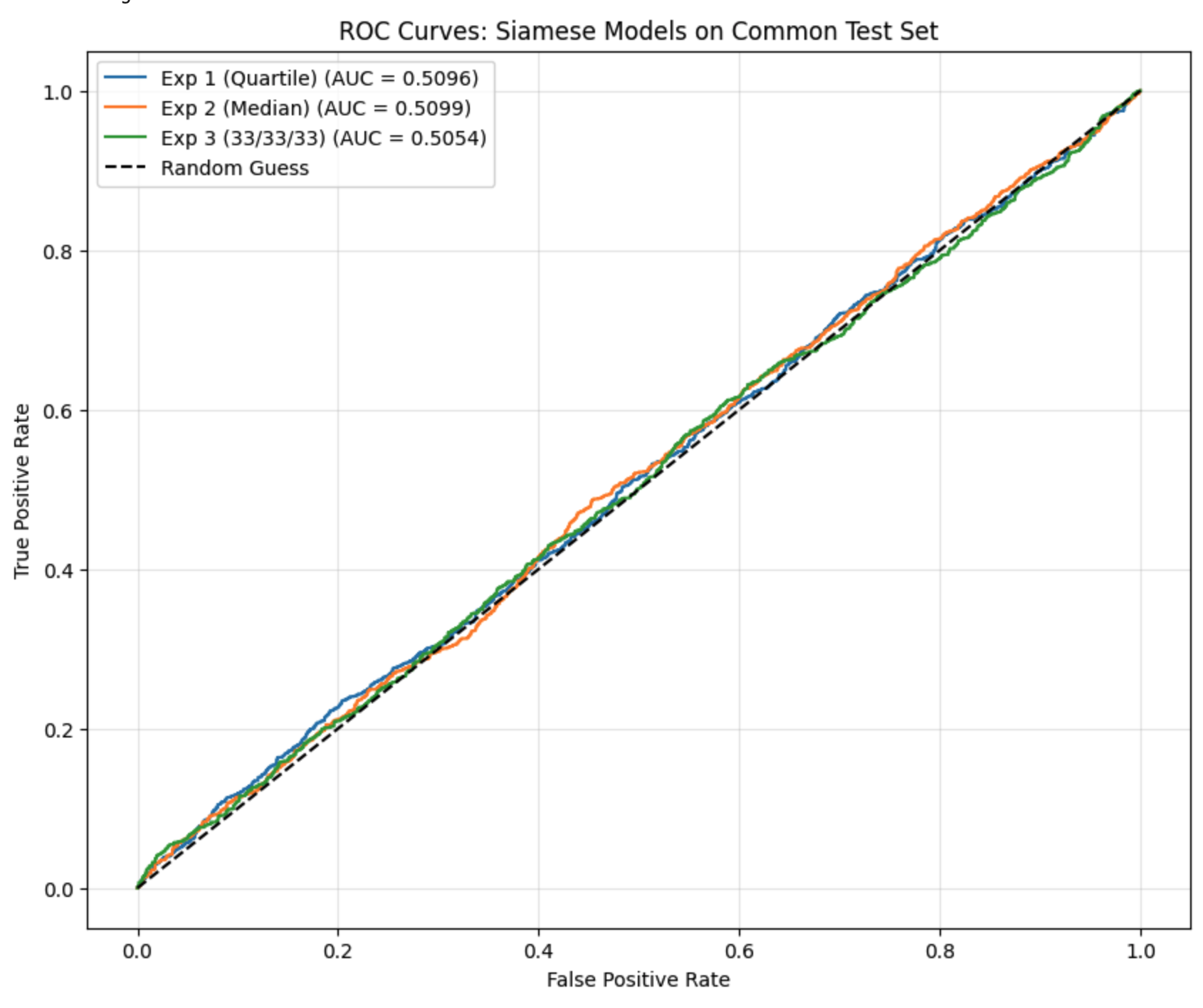}
    \caption{Comparison of ROC curves across Siamese networks for binning strategies}
    \label{fig:siamese_roc}
\end{figure}

\begin{figure}[htbp]
    \centering
    \includegraphics[width=1.0\linewidth]{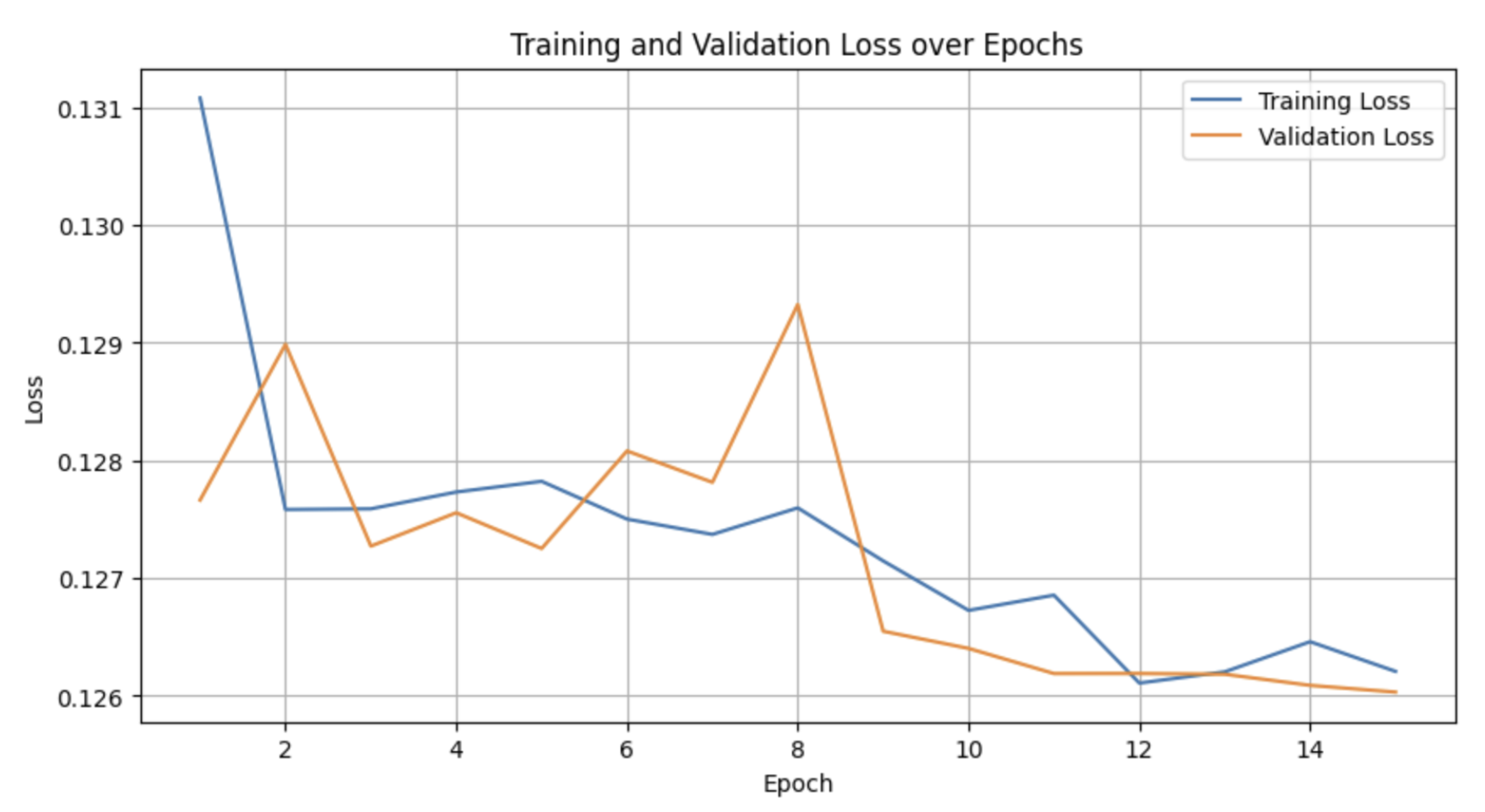}
    \caption{Loss of Siamese network (Quartile Strategy)}
    \label{fig:lossq}
\end{figure}

\begin{figure}[htbp]
    \centering
    \includegraphics[width=1.0\linewidth]{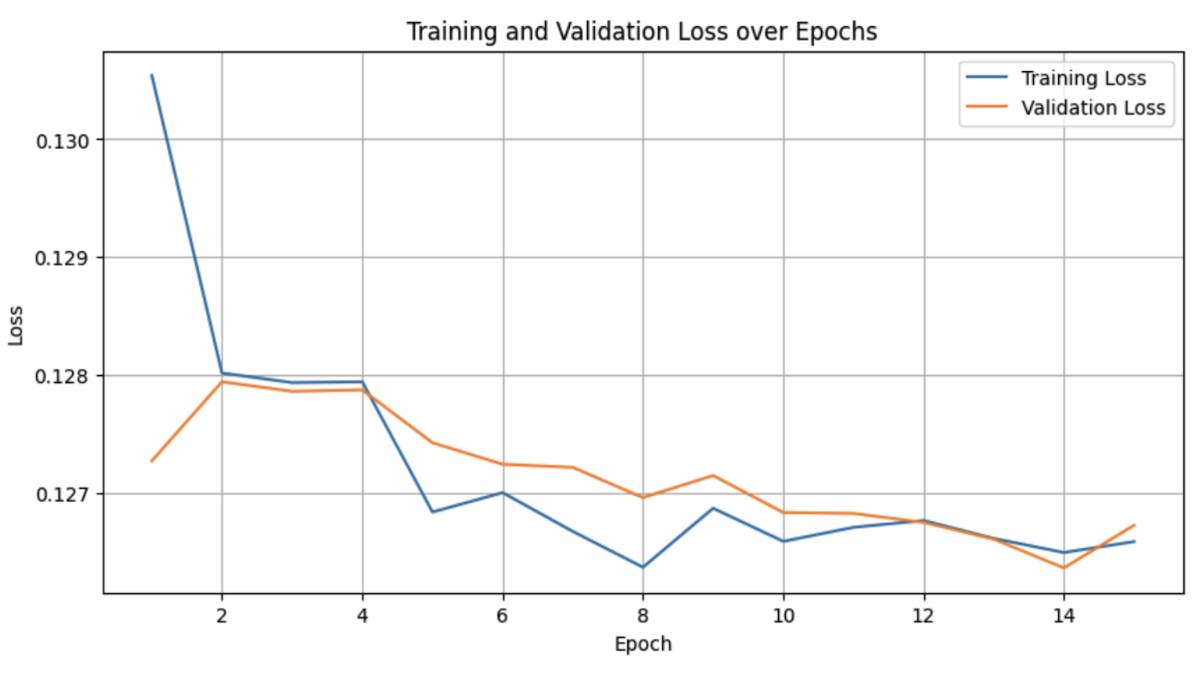}
    \caption{Loss of Siamese network (Median Strategy)}
    \label{fig:lossm}
\end{figure}

\begin{figure}[htbp]
    \centering
    \includegraphics[width=1.0\linewidth]{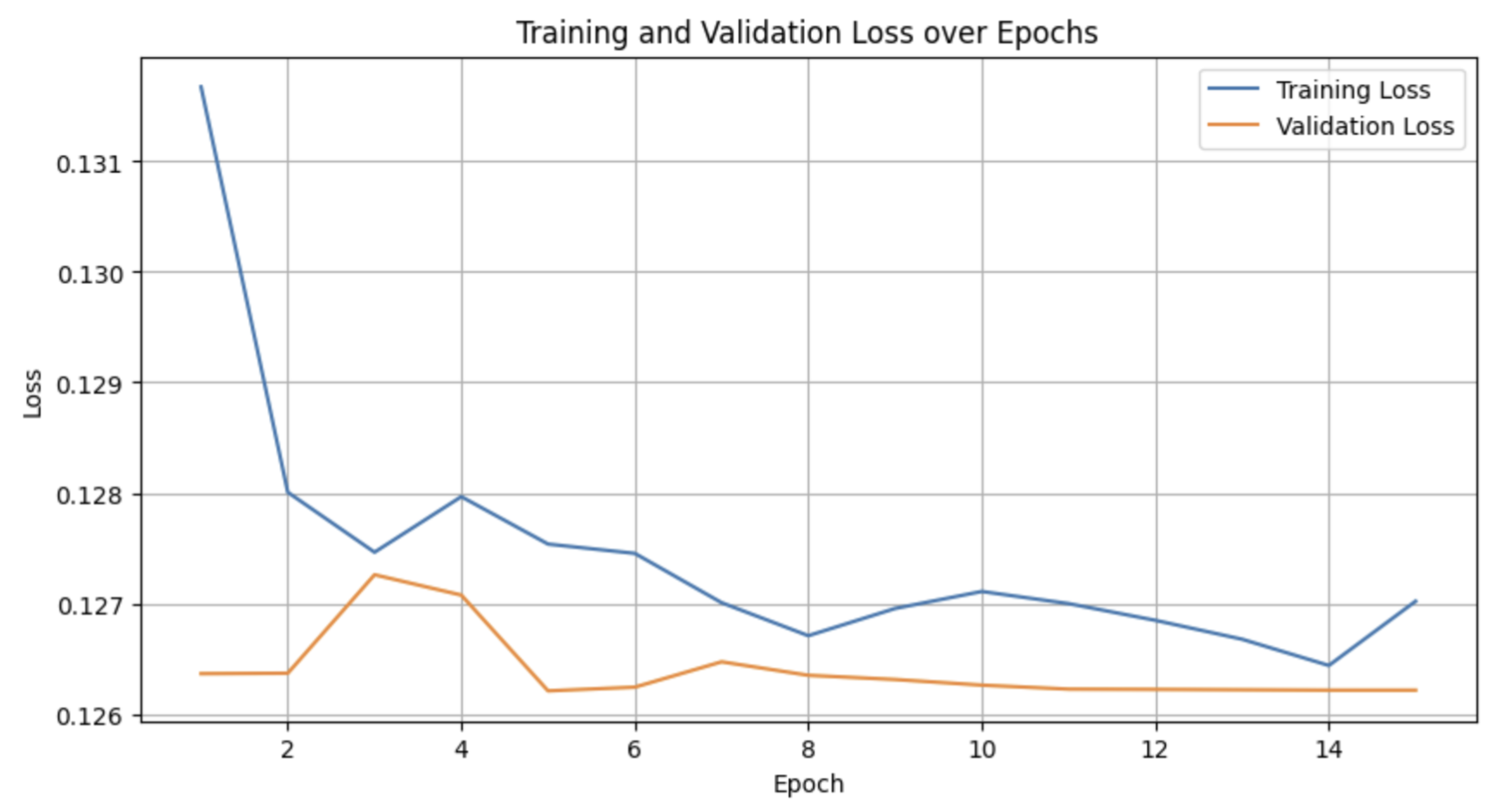}
    \caption{Loss of Siamese network (Tercile Strategy)}
    \label{fig:losst}
\end{figure}

\end{document}